\PassOptionsToPackage{ruled, vlined, linesnumbered}{algorithm2e}
\documentclass{l4dc2024}

\theorembodyfont{\normalfont}
\newtheorem{proposition}{Proposition}
\newtheorem{definition}{Definition}

\newtheorem{corollary}{Corollary}
\newtheorem{lemma}{Lemma}
\newtheorem{theorem}{Theorem}
\newtheorem{remark}{Remark}
\newtheorem{assumption}{Assumption}

\usepackage{cases}
\usepackage{amsfonts}
\usepackage{mathtools} %% for \coloneqq
\usepackage{mathrsfs}
\usepackage{nicefrac}
\usepackage{cite}
\usepackage{graphicx}
\usepackage{adjustbox}
\usepackage[font=footnotesize,labelfont=bf]{caption}
\usepackage{subcaption}

\usepackage{multirow}
\usepackage{bm}
\usepackage{soul}
\newcommand{\R}{\mathbb{R}}

\newcommand{\traj}{\mathrm{x}}
\newcommand{\ctrl}{\mathrm{u}}

\newcommand{\cfset}{\mathcal{U}}

\newcommand{\uref}{\pi_{\text{ref}}}
\newcommand{\usafe}{\pi_{\text{safe}}}
\newcommand{\cbf}{h}
\newcommand{\dcbf}{\frac{\partial \cbf}{\partial x}}

\title[Learning Discriminating Hyperplanes]{Safety Filters for Black-Box Dynamical Systems by Learning Discriminating Hyperplanes}
\usepackage{times}

\author{%
 \Name{Will Lavanakul*} \Email{lav.will@berkeley.edu}\\
 \addr University of California, Berkeley
 \AND
 \Name{Jason J. Choi*} \Email{jason.choi@berkeley.edu}\\
 \addr University of California, Berkeley
 \AND
 \Name{Koushil Sreenath} \Email{koushils@berkeley.edu}\\
 \addr University of California, Berkeley
 \AND
 \Name{Claire J. Tomlin} \Email{tomlin@eecs.berkeley.edu}\\
 \addr University of California, Berkeley
\thanks{* Indicates co-first authors. Extended version.} % Extended version with appendix is available at \citet{lavanakul}.}
}

\begin{document}

\maketitle
\vspace{-3em}
\begin{abstract}%
Learning-based approaches are emerging as an effective approach for safety filters for black-box dynamical systems. Existing methods have relied on certificate functions like Control Barrier Functions (CBFs) and Hamilton-Jacobi (HJ) reachability value functions. The primary motivation for our work is the recognition that, ultimately, enforcing the safety constraint as a control input constraint at each state is what matters. By focusing on this constraint, we can eliminate dependence on any specific certificate function-based design. To achieve this, we define a \textit{discriminating hyperplane} that shapes the half-space constraint on control input at each state, serving as a sufficient condition for safety. This concept not only generalizes over traditional safety methods but also simplifies safety filter design by eliminating dependence on specific certificate functions. We present two strategies to learn the discriminating hyperplane: (a) a supervised learning approach, using pre-verified control invariant sets for labeling, and (b) a reinforcement learning (RL) approach, which does not require such labels. The main advantage of our method, unlike conventional safe RL approaches, is the separation of performance and safety. This offers a reusable safety filter for learning new tasks, avoiding the need to retrain from scratch. As such, we believe that the new notion of the discriminating hyperplane offers a more generalizable direction towards designing safety filters, encompassing and extending existing certificate-function-based or safe RL methodologies.
\end{abstract}


\section{Introduction}

While learning-based control has demonstrated capabilities in solving complex tasks for uncertain dynamical systems, safety assurance remains an unresolved challenge. Safe Reinforcement Learning (RL) has made strides towards co-learning performance and safety \citep{garcia2015comprehensive}, but the absence of provable guarantees in these data-driven methods remains a significant challenge. Conversely, although model-based control theory has well-established mechanisms to enforce safety constraints, its application to uncertain systems is limited when the models deviate from the actual system dynamics.

In an effort to bridge this gap and provide safety guarantees for black-box dynamical systems with unknown closed-form expressions, researchers are actively combining model-based control techniques with data-driven methods \citep{brunke2022safe, Wabersich2023}. Traditional model-based methods for safety problems fundamentally involve two key components: a safe set where the trajectory can remain indefinitely, and a control input constraint that guarantees this invariance within the set. These methods typically rely on certificate functions such as Lyapunov functions, Control Barrier Functions \citep{Ames2016}, or HJ reachability value functions \citep{Fisac2018}. Consequently, existing learning-based methods for designing safety filters for black-box systems have primarily focused on learning the certificate functions \citep{dawson23tro}.

We endorse the philosophy that control theory-based analysis can play a crucial role in devising a data-driven scheme for safety control. However, the main motivation for our work comes from the realization that the enforcement of an appropriate control input constraint at each state is the ultimate step for guaranteeing safety. By focusing on the constraint, we eliminate the need to rely on a specific certificate function-based safe set representation and the input constraint form.

Towards this end, we define a \textit{discriminating hyperplane} that shapes the half-space constraint on control input at each state and serves as a sufficient condition to ensure control invariance of a safe set. The discriminating hyperplane-based safety constraint generalizes the well-known Nagumo-like condition \citep{blanchini1999set}, CBF constraint \citep{Ames2016}, and HJ reachability optimal control law 
\citep{Fisac2018}, for control-affine systems.

Next, we present two approaches for learning the discriminating hyperplane for black-box systems. The first method utilizes supervised learning, where the learning is guided by labels derived from a pre-verified control invariant set. Although this assumes access to prior knowledge of the invariant set, it can be well supplemented by leveraging existing research focused on constructing or learning control invariant sets \citep{bertsekas1972infinite, blanchini1999set, Fisac2018, wabersich2018scalable, bansal2021deepreach, cao2022efficient}. In our simulations across various systems, we find that the safety filter, based on the learned discriminating hyperplane, closely approximates a filter that could be designed using ground-truth system dynamics.

The second method, utilizing reinforcement learning (RL), circumvents the need for access to control invariant sets by learning the discriminating hyperplane directly from trajectory data. We can employ existing RL algorithms, such as Proximal Policy Optimization (PPO) \citep{schulman2017ppo}, for this purpose. A key strength of this approach is the ability to separate a learning-based policy into two components--task achievement and safety filtering. This separation allows the learned safety filter to be reused or adapted to various performance tasks. The concept of separating performance and safety is well-supported by existing safe RL literature \citep{thananjeyan2021recovery, wagener2021safe, kim2023not}. The main distinction of our work is that there is no on-off switching mechanism between performance and safety policies. Instead, we integrate these two aspects into a unified safety filter design based on the discriminating hyperplane concept.

\section{Related Work}
\label{sec:related_work}

\textbf{Certificate functions:} Many model-based control approaches for safety problems use the concept of certificate functions \citep{Prajna2006, dawson23tro}. These are also referred to as safety index in \citet{liu2014control} or energy function in \citet{wei2019safe}. Put simply, a certificate function is a state-dependent scalar function, whose level sets characterize the safe domain and whose gradient can impose a constraint on the control input to ensure safety. To address difficulties in designing certificate functions using classical methods, learning-based methods are being actively explored \citep{fisac2019bridging, srinivasan2020learning, huh2020safe, lindemann2021learning, thananjeyan2021recovery, liu2023safe, so2023train, castaneda2023distribution}.

\noindent \textbf{Learning constraints for uncertain systems:} A crucial step in designing the safety filter is to come up with a valid constraint to be imposed on the control input. Usually, this constraint is a sufficient condition for the closed-loop trajectory to stay invariant inside the safe domain. These constraints can be derived from certificate functions. For uncertain systems, these are typically designed based on a nominal model, and are combined with data-driven methods to accommodate the modeling error \citep{choi2020reinforcement, castaneda2021pointwise, taylor2020cbf, taylor2020towards, brunke2022safe, Wabersich2023}. % When access to a nominal model is not possible, these approaches necessitate the concurrent learning of the certificate functions \citep{qin2022sablas}.
Our methodology aligns closely with these techniques. However, crucially, our approach is independent of both certificate functions and nominal models.

\noindent \textbf{Safe Reinforcement Learning (RL):} Numerous safe RL algorithms proposed in \citet{garcia2015comprehensive, pmlr-v70-achiam17a, ray2019benchmarking, srinivasan2020learning, thananjeyan2021recovery, wagener2021safe} address safety using a purely data-driven approach. A majority of these methods construe safety violations as an \textit{accumulation} of safety-relevant cost, referred to as the constrained Markov Decision Process problem \citep{Altman1998}. While such a formulation finds applicability in certain safety contexts, it falls short in capturing instantaneous constraint violations like collision. While some safe RL algorithms incorporate control theory concepts---for example, Lyapunov theory in \citet{chow2018lyapunov} and reachability theory in \citet{fisac2019bridging, huh2020safe, hsu2023isaacs}---these ideas are applied in the algorithmic design rather than in the structural design of the learned safe policy. In contrast, our approach imposes a structural knowledge derived from the discriminating hyperplane to the learned safety filter. Approaches in \citet{cheng2019end, emam2022safe} explicitly use certificate functions to achieve this. % Other distinctions between safe RL and our method are discussed in Remarks \ref{remark:saferl1}.

\section{Discriminating Hyperplane}
\label{sec:preliminaries}

\subsection{Safety filters for black-box dynamical systems}

We are interested in guaranteeing safety for a black-box dynamical system, which is characterized by a state trajectory, $\traj(t)$, a solution to an ordinary differential equation (ODE) with an initial condition $\traj(0) = x$. % This formulation is a common means of characterizing a deterministic dynamical system that evolves continuously over time \citep{sastry2013nonlinear}. 
Specifically, we make the following assumptions about the system dynamics.
\vspace{-0.25em}
\begin{assumption}[System dynamics]
\label{assump:dynamics} The dynamics is \textit{affine in control}, represented by an ODE
    \vspace{-0.25em}
\begin{align}
    \dot{\traj}(t) = f(\traj(t)) + g(\traj(t))\ctrl(t)\;\;\text{for}\;\;t>0, \qquad \traj(0)=x,
    \label{eq:dynamics}    
\vspace{-0.25em}
\end{align}    
where $x\in \R^n$ is an initial state, $\traj:[0,\infty) \rightarrow \R^n$ is the solution to the ODE, and $\ctrl:[0,\infty)\rightarrow U \subset \R^m$ is a control signal. $f\!:\!\R^n \!\rightarrow\!\R^n$, $g\!:\!\R^n\!\rightarrow\!\R^{n \times m}$ are bounded Lipschitz continuous vector fields. The control input set $U$ is compact.
\vspace{-0.25em}
% \begin{enumerate}
%     \item The dynamics is \textit{affine in control}, represented by an ODE
%     \vspace{-0.25em}
% \begin{align}
%     \dot{\traj}(t) = f(\traj(t)) + g(\traj(t))\ctrl(t)\;\;\text{for}\;\;t>0, \qquad \traj(0)=x,
%     \label{eq:dynamics}    
% \end{align}    
% where $x\in \R^n$ is an initial state, $\traj:[0,\infty) \rightarrow \R^n$ is the solution to the ODE, and $\ctrl:[0,\infty)\rightarrow U \subset \R^m$ is a control signal.
% \vspace{-0.5em}
% \item $f\!:\!\R^n \!\rightarrow\!\R^n$, $g\!:\!\R^n\!\rightarrow\!\R^{n \times m}$ are \textit{bounded Lipschitz continuous} vector fields.
% \vspace{-0.5em}
% \item \textit{Bounded control: } The control input set $U$ is compact.
% \vspace{-0.5em}
% \end{enumerate}
\end{assumption}

These assumptions reflect the realistic conditions of real-world physical systems and set the minimum requirements for the structure of the systems under consideration. The control-affine nature of the dynamics applies to a variety of physical systems derived from Euler-Lagrangian mechanics, or the dynamics can often be transformed into a control-affine form through a change of variables. The last condition is typical of physical systems with bounded actuation limits.

In our framework, we do not assume access to the closed-form expressions of $f$ and $g$. Instead, we assume the ability to sample state trajectories over discrete steps in time, through simulation or experiments. With this setup, we can efficiently use the collected state transition data to train the safety filter without requiring explicit knowledge of the underlying system dynamics.

The safety problem we focus on is to ensure that the system states satisfy specific constraints over an indefinite time horizon. Thus, safety for the system \eqref{eq:dynamics} is encoded by a \textit{target constraint set} $X \subset \R^{n}$ that must be respected during the evolution of the system:
\vspace{-0.5em}
\begin{equation}
    \traj(t)\in X ~ \text{for all}~t\ge 0.
  \label{eq:constraints}
\vspace{-0.5em}
\end{equation}
\noindent The \textit{safety filter} we aim to design operates as an intermediary between the reference controller $\uref:\R^n\rightarrow U$ and the system, ensuring that the downstream control adheres to the safety constraint described in \eqref{eq:constraints}. Reference controllers are typically safety-agnostic and can take various forms, from a neural network policy optimized for performance objectives, hand-designed controllers from domain experts, to a human operator's commands. 
The safety filter $\usafe: \R^n \rightarrow U$ modifies the input signal from the reference controller when necessary, yielding a control input signal $\ctrl(t) = \usafe(\traj(t); \uref)$. We mainly focus on the \textit{safety constraint} within the filter, which modifies the control input to guarantee safety:
\begin{definition}[Safety constraint \& safe domain for safety filter] \label{def:cert}
We say that $c(x, u) \ge 0$ is a (valid) \textit{safety constraint} if there exists a \textit{safe domain} $S \subseteq X$, such that for all Lipschitz feedback policy $\pi:\R^n \rightarrow U$ that satisfies $c(x, \pi(x)) \ge 0$ for all $x\in S$, the trajectory resulting from the control signal $\ctrl(t) = \pi(\traj(t))$ satisfies \eqref{eq:constraints} for all $x \in S$ where $\traj(0) =  x$.
\end{definition}
Note that the Lipschitz continuity of $\pi$ is required to guarantee the solution $\traj(\cdot)$ exists and is unique. In words, the safety constraint is a sufficient condition that, when $u$ satisfies the constraint, safety is guaranteed for all states in the safe domain $S$, a subset of the target constraint set $X$. Consequently, the effective design of a safety filter hinges on the choice of a valid safe domain $S$ and a safety constraint $c(x, u) \ge 0$.

We first review how a safe domain is generally designed in the literature. Many existing design approaches \citep{Wabersich2023} seek to find a set $S$ that is control invariant, defined next.

\begin{definition}[Control Invariance \citep{blanchini1999set}]
\label{def:ci}
A set $S \subset \R^n$ in the state space is \textit{control invariant} (under the dynamics \eqref{eq:dynamics}) if for all $x \in S$, there exists a control signal $\ctrl\in\cfset$ such that 
\vspace{-0.5em}
\begin{equation}
    \traj(t)\in S ~ \text{for all}~t \ge 0.
\vspace{-0.5em}    
  \label{eq:ci-constraint}
\end{equation}
\end{definition} 
The definition implies that a control invariant set $S$ that is a subset of $X$ can serve as a safe domain. The representation of the control invariant set can vary significantly depending on the chosen design methodology. It ranges from geometric structures like polytopes \citep{blanchini1999set} to level sets of scalar functions such as certificate functions \citep{dawson23tro}, and extends to the concept of feasibility in receding-horizon optimal control \citep{wabersich2021predictive}. This versatility underscores control invariance as a foundational concept in designing safe domains.

\subsection{Discriminating hyperplane for safety constraint}
\label{sec:our-theory}

Once a control invariant safe  domain $S$ is specified, a similar blueprint for characterizing a valid safety constraint exists. It is based on the geometric relationship that the control invariant set and the vector field of the dynamics satisfy at its boundary, $\partial S$, which is known as the ``Nagumo-like'' condition in the literature \citep{nagumo1942lage, blanchini1999set, aubin2011viability, Ames2016}:
\begin{lemma}
\label{lemma:tangential_ci} (Tangential characterization of control invariant sets \citet[Theorem 11.3.4]{aubin2011viability}) Let the dynamics \eqref{eq:dynamics} satisfy Assumption \ref{assump:dynamics}. Then, a closed set $S \subset \R^n$ is control invariant if and only if for all $x\in \partial S$,
\vspace{-0.5em}
\begin{equation}
\label{eq:ci_tangent}
    \exists u \in U ~\textnormal{such that}~ \dcbf(x) \cdot (f(x) + g(x) u) \ge 0,
\end{equation}
where $\cbf\!:\!\R^n \rightarrow \R$ is continuously differentiable function such that $S\!=\!\{x| h(x) \ge 0\}$ and $\dcbf(x)$ is bounded away from 0 for all $x \in \partial S$, implying that the slope of $h$ does not vanish at the boundary\footnote{Such $\cbf$ exists for $S$ whose interior is not empty and boundary is continuously differentiable \citep{lieberman1985regularized}.}.
\end{lemma}

The condition \eqref{eq:ci_tangent} implies that for the set to be control invariant, there must exist a control input that renders the vector field pointing inward to the set. In fact, the condition is a geometric property that remains invariant under different choices of the distance-like function $\cbf$, and \eqref{eq:ci_tangent} is merely its analytic description\footnote{\eqref{eq:ci_tangent} can be rewritten as $\exists u\!\in\!U$ s.t. $(f(x) \!+\!g(x)u)\!\in\! T_S(x)$, where $T_S$ is (Bouligand's) tangent cone to $S$ \citep{clarke2008nonsmooth}.}. Extending this property, in the new notion of the discriminating hyperplane we introduce next, the resulting safety constraint also does not rely on any choice of $h$. In contrast, the notion of CBF in \citep{Ames2016}, which is also inspired by Lemma \ref{lemma:tangential_ci}, results in a certifying constraint that is dependent on $h$.

From \eqref{eq:ci_tangent}, a constraint which is affine in $u$, imposed whenever the state $x$ is at the boundary of $S$, is sufficient to regulate the trajectory to not exit the set $S$. This serves as a starting point for constructing a control-affine constraint extended to all $x \in S$ including the interior of the set. By noticing that the control-affine inequality defines a halfspace in the control input space, we define the discriminating hyperplane as below:
\begin{definition}
\label{def:dh}
    A \textit{discriminating hyperplane} for systems satisfying Assumption \ref{assump:dynamics} and a control invariant set $S \in \R^n$ is a hyperplane defined in the control input space represented by $a(x)^\top u = b(x)$, for each state $x \in S$, such that the resulting half-space constraint $a(x)^\top u \ge b(x)$ is a safety constraint according to Definition \ref{def:cert}.
\end{definition}
In words, the discriminating hyperplane distinguishes between certified input signals that guarantee safety, and uncertified input signals that could potentially lead to safety violations. The hyperplane is defined in the control input space and is parameterized by each state. If such a discriminating hyperplane can be determined, we can use it effectively to construct the following safety filter:
\vspace{0.5em}
\hrule
\vspace{0.5em}
\noindent \textbf{Discriminating hyperplane-based min-norm safety filter}:
\begin{equation}
\begin{aligned}
\label{eq:dh-qp}
\usafe(x)=\arg\min_{u \in U} \quad & ||u-\uref(x)||^2\\
\textrm{s.t.} \quad & a(x)^\top u \geq b(x)\\
\end{aligned}
\end{equation}
\hrule
\vspace{0.5em}
Note that this filter selects a control input that is certified to be safe by the discriminating hyperplane and is closest to the reference control input $\uref(x)$. The filter becomes a quadratic program when $U$ is a polytope. When $U$ is a general convex set, the program is still a convex program.

Next, we present a sufficient condition for $a(x)^\top u = b(x)$ to be a discriminating hyperplane. In words, the theorem says that any discriminating hyperplane leading to the satisfaction of the Nagumo-like condition \eqref{eq:ci_tangent} is valid.
\begin{theorem}
\label{thm:main1}
For the control input set $U$ that is polytopic, if $a:\R^n\rightarrow\R^{m}$ and $b:\R^n\rightarrow \R$ are Lipschitz continuous in $x$, and if for all $x \in \partial S$, $\{u \in U \;|\;a(x)^\top u \geq b(x)\}$ has a nonempty interior and is a subset of $\{u \in U \;|\;\dcbf(x) \cdot (f(x) + g(x) u) \ge 0\}$, then $a$ and $b$ define a discriminating hyperplane, $\{u \;| \;a(x)^\top u = b(x)\}$. % (Proof: \citet[Appendix]{lavanakul}) 
(Proof: Appendix \ref{subsec:proof1})
\end{theorem}
An important lemma for the theorem's proof establishes that $\usafe$ in \eqref{eq:dh-qp} is the feedback policy that shows the validity of $a(x)^\top u \ge b(x)$ as a safety constraint satisfying Definition \ref{def:cert}:

\begin{lemma}
\label{cor:theorem1}
If $\uref:\R^n\rightarrow U$ is Lipschitz continuous in $x$ and if $a, b$ satisfy the conditions in Theorem \ref{thm:main1}, then $\usafe$ in \eqref{eq:dh-qp} is also Lipschitz continuous in $x$.
\end{lemma}

\begin{remark}
CBFs in \citep{Ames2016} and HJ reachability value functions in \citep{Fisac2018} offer special cases of the discriminating hyperplane. For a CBF $h$, the hyperplane is given by $a(x)=\dcbf(x) \cdot g(x)$ and $b(x)=-\dcbf(x) \cdot f(x)-\alpha(h(x))$, where $\alpha$ is the comparison function associated with the CBF. For a reachability value function $V$, the hyperplane is given by $a(x)=\frac{\partial V}{\partial x}(x) \cdot g(x)$ and $b(x)=-\frac{\partial V}{\partial x}(x) \cdot f(x)$. As such, the discriminating hyperplane can be considered as a generalized structure of safety constraints for control-affine systems that unifies and extends the results of the existing methods.
\end{remark}

\subsection{Sample-and-hold lookahead-based discriminating hyperplane}
\label{sec:lookahead}

We present a simple way to construct a discriminating hyperplane for a given control invariant set $S$, without having to rely on any specific distance-like function $h$ of the set $S$. This is consistent with the core principle of our approach, which is to avoid reliance on any specific form of certificate functions or safe domain representation. Thus, we only rely on a minimal knowledge of the invariant set: an indicator of whether a given state is within $S$: $I_{S}(x) = 1$ if $x\in S$, and $0$ otherwise.

The method we propose is to construct the discriminating hyperplane by checking whether the state trajectory exits the set $S$ for a lookahead time $\Delta t$, under sample-and-hold of control input $u$. This approach is endorsed by the following theorem:
\begin{theorem}
\label{thm:main2}
For $\Delta t > 0$, there exists $a:\R^n\rightarrow\R^{m}$ and $b:\R^n\rightarrow \R$ that are Lipschitz continuous in $x$, such that $\Pi(x) = \{u \in U \;|\;a(x)^\top u \geq b(x)\} \subseteq \{u \in U \;|\;I_S(\traj(\Delta t)) = 1\}$ where $\traj(\cdot)$ is from \eqref{eq:dynamics} with $\traj(0)=x, \ctrl(\cdot)\equiv u$. Moreover, for small enough $\Delta t$, $\Pi(x)$ is not empty if $\{u \in U \;|\;I_S(\traj(\Delta t)) = 1\}$ has a non-empty interior. % (Proof: \citet[Appendix]{lavanakul}) 
(Proof: Appendix \ref{subsec:proof2})
\end{theorem}
\noindent By using the lookahead approach, $\Pi(x)$ satisfying Theorem \ref{thm:main2} anticipates into the future, which is also the main mechanism of CBFs \citep{choi2023forward} or predictive filter \citep{wabersich2021predictive}. If the lookahead time is small, then the constraint would be active only on states that are close to the boundary. In contrast, larger $\Delta t$ will encode additional safety margin from the boundary of $S$, however, it might make $\Pi(x)$ an empty set. In Section \ref{sec:results}, we demonstrate how this design results in a ``smooth braking'' behavior as $\traj(\cdot)$ approaches the boundary of the safe domain, similarly to the CBF-based design, and how different choice of the lookahead time affects this behavior.

\section{Learning Discriminating Hyperplane}
\label{sec:method}

This section presents supervised and reinforcement learning approaches to learn the discriminating hyperplane from black-box system trajectory data.

\subsection{Supervised learning approach}
\label{sec:supervised}
In our supervised learning approach, we use the lookahead-based discriminating hyperplane in Section \ref{sec:lookahead} to construct a training label for the neural network that learns the hyperplane. We train $[a_{\theta}(x), b_{\theta}(x)] \in \mathbb{R}^{m+1}$, where $\theta$ is the neural network weights, in a supervised manner using a dataset of state and input transition pairs, and their labels detailing if an input is safe. Specifically, $N$ states, $\{x_i\}_{i=1}^{N}$, are uniformly sampled from the control invariant set $S$. Then $M$ inputs per state, sampled uniformly from $U$, $\{u_{ij}\}_{j=1}^{M}$, are applied to the dynamical system by sample-and-hold for the lookahead time $\Delta t$, resulting in the terminal state $\traj_{ij}(\Delta t)$. The indicator function of the set $S$ at the terminal state, $I_S(\traj_{ij}(\Delta t))$, determines the label of whether the next state is safe or unsafe:
\vspace{-0.5em}

\small
\begin{equation}
y_{ij} = 
\begin{cases}
    1 & \text{if } I_S\left(\traj_{ij}(\Delta t)\right) = 1 \text{ where } \traj_{ij} \text{ solves \eqref{eq:dynamics} for } \traj_{ij}(0)\!=\!x_i, \ctrl\!\equiv\!u_{ij},\\
    -1 & \text{else.}
\end{cases}
\vspace{-0.25em}
\end{equation}

\normalsize
\noindent In essence, we are labeling the inputs based on whether they leave $S$ after the lookahead time. 

In order for the neural network hyperplane to be a valid discriminating hyperplane, any control input that is predicted to be safe, i.e. $a_\theta(x_i)^\top u_{ij} \geq b_\theta(x_i)$, should have the label $y_{ij} = 1$. As such, we aim to minimize the misclassification rate of the neural network hyperplane, which can be achieved by minimizing the following loss function: \vspace{-0.5em}

\footnotesize
\begin{equation*}
    \mathcal{L} = \frac{1}{N}\sum_{i=1}^N  \left( \gamma_{\text{pos}}\!\!\!\!\!\!\!\!\!\!\sum_{\substack{j: \\ a_{\theta}(x_i)^\top \!\!u_{ij} > b_{\theta}(x_i)}} \!\!\!\!\!\!\!\!\!\!\!\!\!\!\!-\!\min\!\left\{y_{ij}(a_{\theta}(x_i)^\top\!\!u_{ij} - b_{\theta}(x_i)), 0\right\} + \gamma_{\text{neg}}\!\!\!\!\!\!\!\!\!\!\!\sum_{\substack{j:\\a_{\theta}(x_i)^\top 
    \!\!u_{ij} < b_{\theta}(x_i)}}\!\!\!\!\!\!\!\!\!\!\!\!\!\!\!-\!\min\!\left\{y_{ij}(a_{\theta}(x_i)^\top \!\!u_{ij} - b_{\theta}(x_i)), 0\right\}\right).
\end{equation*}

\normalsize
\noindent If the prediction of safety from the neural network matches the labels, the loss is not incurred. False positive samples whose unsafe control input is predicted to be safe are counted in the first term, and false negative samples whose safe control input is predicted to be unsafe are counted in the second term, where $\gamma_{\text{pos}}$ and $\gamma_{\text{neg}}$ are weights for false positive and negative samples. We typically set $\gamma_{\text{pos}} > \gamma_{\text{neg}}$ as it is crucial to rule out unsafe inputs correctly. % It is worth noting that the structure of the loss exhibits similarities with the loss design employed in \citet{dawson23tro} for learning certificate functions.

\subsection{Reinforcement learning approach}
\label{sec:rl}
An alternative approach is to employ RL, when it is hard to find a control invariant set $S$ for the supervised learning method. We propose a method analogous to PPO, where instead of learning a policy that maximizes reward, we use an actor parametrizing the discriminating hyperplane to minimize instances of safety violation. Our method uses the form $\pi^a_{\theta}(a | x), \pi^b_{\theta}(b | x)$ as two normal distributions to sample the hyperplane parameters. %, similar to how PPO samples actions for continuous action spaces. 
During the rollout, based on the sampled $(a, b)$ for each $x$, with probability $1\!-\!\delta$, we sample $u$ from $\{u\in U|\; a^\top u \ge b\}$ which enforces the learned safety constraint, and with probability $\delta$, we sample $u$ from $U$, allowing an exploration with a small probability.
Given the target constraint set $X$, the reward for learning the hyperplane is designed as 
\vspace{-0.75em}

\small
\begin{equation}
\label{eq:rl-dh-reward}
r(x, u)=
    \begin{cases}
        1 + d(u) & \text{if } x' \in X, \\
        c & \text{otherwise,}
    \end{cases}
\end{equation}
\normalsize
\noindent where $x'$ is the next state after taking action $u$, $c \leq 0$ induces a negative reward when safety is violated, as $x' \not \in X$, and $d(u) \geq 0$ is an optional bonus term that can be added to the reward if the input $u$ is in the action space $U$. The bonus term can be necessary when the action space is small, which leads to the actor's predicted hyerplanes being outside of the action space. The reward is positive when safety is satisfied and the bonus term incentivizes the hyperplane to only constrain the input when needed. Let $r^{a}(\theta) = \frac{\pi^a_{\theta}(a | x)}{\pi^a_{\theta_{\text{old}}}(a | x)}$ and $r^{b}(\theta) = \frac{\pi^b_{\theta}(b | x)}{\pi^b_{\theta_{\text{old}}}(b | x)}$, where $\theta_{\text{old}}$ is the policy parameter before the update, and define the clipped surrogate objective terms, $L^{a}(\theta) = \mathop{\mathbb{E}}[\min (r^{a}(\theta)\hat{A}, \text{clip}(r^{a}(\theta), 1-\epsilon, 1+\epsilon)\hat{A})], L^{b}(\theta) = \mathop{\mathbb{E}}[\min (r^{b}(\theta)\hat{A}$, $\text{clip}(r^{b}(\theta), 1-\epsilon, 1+\epsilon)\hat{A})]$, where $\hat{A}$ is the estimated advantage function.
We aim to maximize the combined objective $L^{a}(\theta) + L^{b}(\theta)$, and perform the policy update as PPO does. By doing so, our policy learns the neural network hyperplane that minimizes long term constraint violation while suppressing the conservativeness of the hyperplane. % Once, the discriminating hyperplane is learned, we are able to utilize it in the safety filter \eqref{eq:dh-qp} to train other downstream tasks while minimizing safety violations.

\section{Experiments}
\label{sec:results}
We demonstrate the safety filter in \eqref{eq:dh-qp} based on the discriminating hyperplane (DH) learned through methods described in Section \ref{sec:method} on various dynamical systems\footnote{Implementation details:  % \citet[Appendix]{lavanakul},
Appendix \ref{subsec:appendix-implementation}, \; Supplementary video: \href{https://youtu.be/70xxoVW8z8s}{https://youtu.be/70xxoVW8z8s}\\ Source code: \href{https://github.com/HJReachability/discriminating-hyperplane.git}{https://github.com/HJReachability/discriminating-hyperplane.git} % Implementation details and source code are provide in \citet[Appendix]{lavanakul}.
}. The proposed methods in the paper are each referred to as \textbf{Oracle DH} (DH based on Theorem \ref{thm:main2} in Section \ref{sec:lookahead}, solved with support vector machine for each state), \textbf{SL-DH} (the supervised learning approach in Section \ref{sec:supervised}), and \textbf{RL-DH} (the RL approach in Section \ref{sec:rl}). 
Through these experiments, we want to highlight
\begin{enumerate}
    \item compatibility of DH with various representations of the safe domain $S$, based on HJ reachability in \citet{xue2023reach}, Lyapunov function, and manual design,
    \vspace{-0.5em}
    \item efficacy of the supervised learning approach in Section \ref{sec:supervised} in approximating the Oracle DH,
    \vspace{-0.5em}
    \item effect of lookahead time $\Delta t$ on the SL-DH, and comparison against CBF-based safety filter.
    \vspace{-0.5em}    
    \item usage of the SL-DH and RL-DH for safe RL of versatile tasks, and comparison against unconstrained PPO \citep{schulman2017ppo}, and PPO-Lagrangian \citep{ray2019benchmarking}.
\end{enumerate}

\begin{figure}[t!]
    \centering
\adjustbox{width=\columnwidth}{    \includegraphics{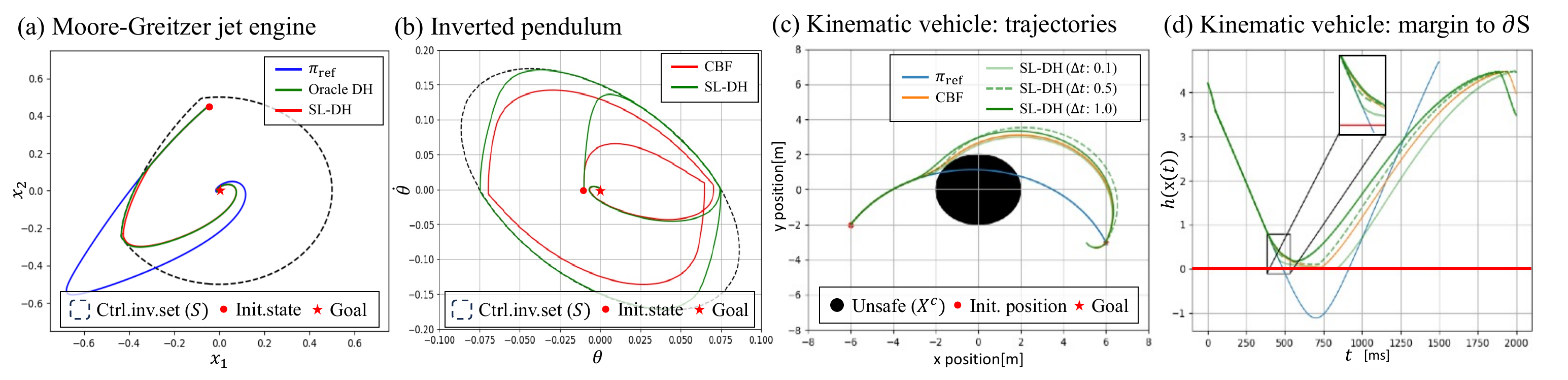}}
\vspace{-2em}
    \caption{Simulation results of safety filter based on SL-DH, using various invariant set representation. (a) Phase plot of the jet engine system in \citet{xue2023reach}, under Oracle DH and SL-DH-based safety filter. The safe set is computed by HJ reachability. (b) Phase plot of the inverted pendulum under a bang-bang reference controller, filtered by CBF \citep{Wabersich2023} and SL-DH. A Lyapunov function level set is chosen as $S$. (c) Position trajectories of the kinematic vehicle, where $S$ is hand-designed. While the reference controller passes through the unsafe region, safety filters based on CBF and SL-DH are able to keep the vehicle safe. In the supplementary \href{https://youtu.be/70xxoVW8z8s}{video}, we provide an animation of the car trajectory under the SL-DH filter using a lookahead time of 0.5. (d) Margin to the boundary of $S$ over time. $h(x)\geq 0$ represents that the state is inside $S$. A larger lookahead time allows the SL-DH to engage in safe actions earlier.}
    \label{fig:exp1}
    \vspace{-1em}
\end{figure}

\noindent \textbf{1. Moore-Greitzer jet engine} (Oracle DH vs. SL-DH, Figure \ref{fig:exp1}(a)): The dynamics of this system is provided in \citet{xue2023reach} ($n$=2, $m$=1), in which the maximal control invariant set of the target constraint set $X=\{x|\sqrt{x_1^2 + x_2^2} \le 0.5\}$ is computed by the discounted infinite-horizon HJ reachability formulation. The noticeable feature of this value function is that it is flat and zero everywhere inside the verified invariant set $S$. Thus, the safety constraint cannot be obtained from the value function, where the DH can be of great use. $\uref$ is designed to stabilize to the origin, which can exit $S$ and violate safety. We use the computed $S$ to train the SL-DH, and compare the closed-loop trajectories under (a) $\uref$, and safety filters based on (b) the Oracle DH, and (c) the SL-DH. The SL-DH safety filter closely approximates the Oracle DH safety filter.

\noindent \textbf{2. Inverted Pendulum} (CBF vs. SL-DH, Figure \ref{fig:exp1}(b)): % The state consists of the angle $\theta$ and the angular velocity $\dot{\theta}$, and the input is the torque ($n$=2, $m$=1). 
The reference controller $\uref$ bounces between extreme torques and turns into a stabilizing controller after 12s. It easily exits the target angle region $X\!=\!\{x|\;|\theta|\!<\!0.3\}$. A level set of a quadratic Lyapunov function in $X$ is chosen as $S$. A CBF $h$ is also derived from this Lyapunov function. The dynamics of this system and $\uref, S, h$ are detailed in \citet{Wabersich2023}. Both CBF and SL-DH filters successfully maintain the system within the safe set while intervening $\uref$ smoothly before the trajectory hits the boundary. This demonstrates how SL-DH is able to achieve similar behavior to CBF without utilization of known system dynamics, assuming access to the safe set beforehand.

\noindent \textbf{3. Kinematic Vehicle} (Effect of $\Delta t$ on SL-DH, Figure \ref{fig:exp1}(c, d)): The state of the vehicle consists of x and y position, heading, and velocity, while the input consists of the yaw rate and acceleration ($n$=4, $m$=2). The reference controller is a feedback controller designed to navigate to a goal point. The control invariant set and the CBF is hand-designed to avoid entering an unsafe region centered at the origin. More details are available in Appendix. % \citet[Appendix]{lavanakul}. 
We train SL-DH with multiple lookahead times: $\Delta t =0.1, 0.5, 1$. While $\uref$ passes through the unsafe region, safety filters based on SL-DH and CBF are able to prevent the vehicle from entering it. A lower lookahead time leads to a less restrictive but more myopic intervention of the filter, resulting in trajectories approaching closer to the unsafe region. With a larger lookahead time, the safety filter is more preemptive with respect to the boundary. We also note that a higher lookahead time can be beneficial for the learning in that the resulting DH varies more smoothly with respect to state. However, an excessively high value can result in the safety filter to be overly conservative.

\begin{figure*}[t!]
    \centering
\adjustbox{width=0.96\columnwidth}{    \includegraphics{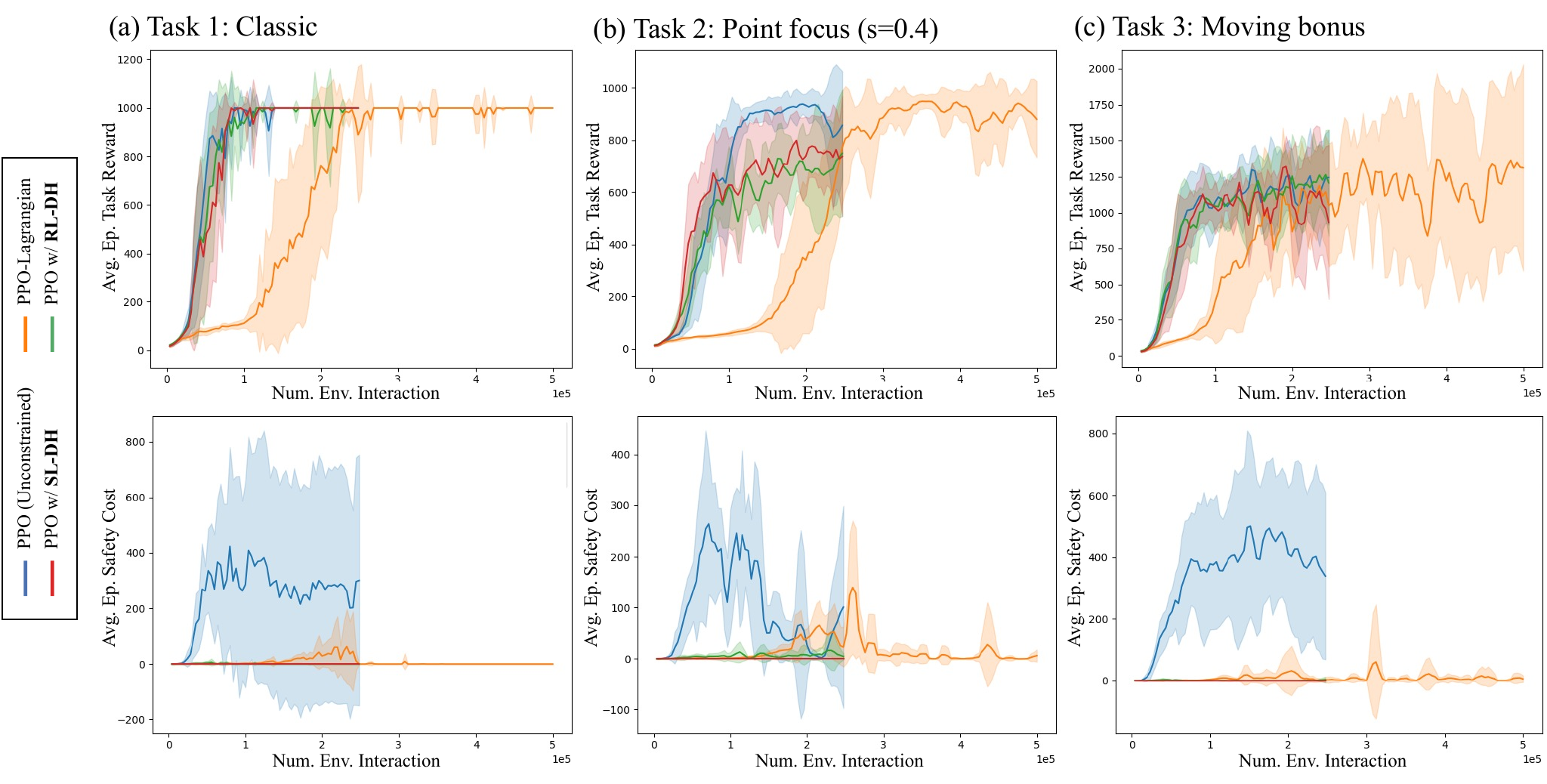}}
\vspace{-1em}    \caption{Cart-Pole Experiment: We compared the reward (top) and constraint cost (bottom) across training iterations of PPO, PPO-Lagrangian, and PPO filtered by SL-DH and RL-DH, across various tasks. Notably, SL-DH and RL-DH did not require retraining for each new task. The policies filtered by these methods resulted in minimal safety violations while effectively learning the task, showing performance comparable to that of the unconstrained PPO in all tasks.}
    \label{fig:cart-pole}
    \vspace{-1em}
\end{figure*}

% \subsection{Cart-pole}
\noindent \textbf{4. Cart-Pole} (Safe RL for various tasks, Figure \ref{fig:cart-pole}): We show the utility of SL-DH and RL-DH for learning various tasks, subjected to the identical target constraint. Specifically, we compare PPO, whose policy is filtered by SL-DH and RL-DH during its task training, to unconstrained PPO and PPO-Lagrangian. The CartPole environment is defined in \citet{towers_gymnasium_2023} ($n=4, m=1$). The considered target constraint is $X\!=\!\{x|\;|s|\leq 0.5\}$, where $s$ is the position of the cart. The classic task is to maximize the length of the trajectory before termination (Task 1). We define two new tasks: stabilizing the cart at $s=0.4$ (Task 2), and maximizing cart speed while avoiding termination (Task 3). These new tasks challenge to approach the constraint's boundary more than the classic task. Training of the SL-DH and RL-DH is detailed in Appendix. % \citet[Appendix]{lavanakul}. 
Both SL-DH and RL-DH filtered PPOs demonstrate competitive performance compared to unconstrained PPO. Moreover, SL-DH shows no constraint violations, while RL-DH exhibits significantly fewer violations than PPO-Lagrangian across all tasks. RL-DH also achieved zero constraint violations in many instances. We also discovered that our method results in lower constraint violations across a broader range of initial states in the state space, as visualized in Appendix.  % \citet[Appendix]{lavanakul}. 
The training of the PPO-Lagrangian is slower and violates safety more compared to our methods, mainly caused by the simultaneous learning of performance and safety, associated with unstable updates of the Lagrange multiplier.

\begin{figure*}[t!]
    \centering
\adjustbox{width=0.95\columnwidth}{    \includegraphics{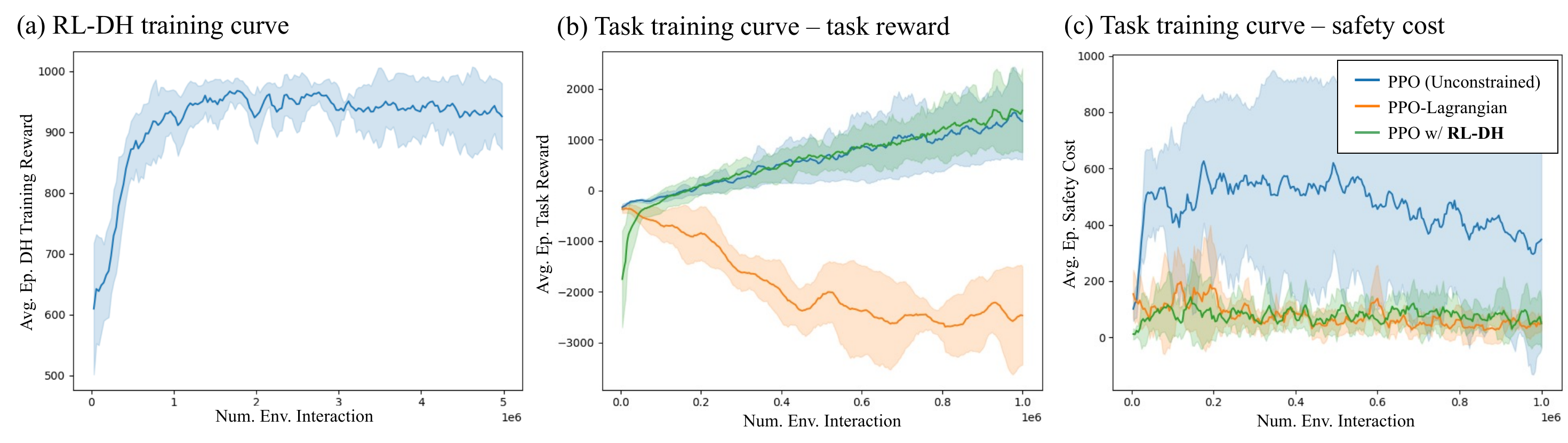}}
    \vspace{-1em}
    \caption{HalfCheetah Experiment: (a) Training curve of RL-DH, where the reward is defined in \eqref{eq:rl-dh-reward}. The value achieving 1000 implies no constraint violation of the trajectory. (b), (c) Reward and constraint cost across iterations of PPO, PPO-Lagrangian, and PPO filtered by RL-DH during the task training.}
    \label{fig:half-cheetah}
    \vspace{-1em}
\end{figure*}

\noindent \textbf{5. HalfCheetah} (Safe RL for uncertain high-dimensional system, Figure \ref{fig:half-cheetah}): We show the viability of RL-DH on a higher dimensional system through HalfCheetah in \citet{towers_gymnasium_2023}. In our setup, we utilize a safety constraint of $X= \{x | z \geq -0.3\}$ where $z$ is the height of the robot torso. This constraint represents the requirement for the robot to stay upright during training. The environment consists of $n\!=\!16$ states and $m\!=\!6$ control variables, whose dynamics are hard to model due to the contact with the ground. Training SL-DH for this example is challenging, as it requires a design of a control invariant set for the complicated dynamics.
PPO filtered by the trained RL-DH maintains performance competitive with that of unconstrained PPO, while achieving a level of constraint violation comparable to PPO-Lagrangian. However, we observe that RL-DH experiences a distribution shift between the training distribution of the hyperplane and that of the task-specific PPO policy. The simulation results of the trained policies can be found in the supplementary \href{https://youtu.be/70xxoVW8z8s}{video}.

\section{Conclusion and Future Work}
\label{sec:conclusion}
In this work, we have proposed a novel learning-based approach for the design of safety filters, focusing primarily on the control input constraint, which is key to ensuring safety. Our method, which employs a neural network to produce parameters of a \textit{discriminating hyperplane}, offers a more general viewpoint of how to construct safety filters for general control-affine systems. We have shown that our approach is modular, working together with any control invariant set representation, and also compatible with any performance-driven objectives, thus replacing the need for safe RL approaches that combines performance and safety. While we have shifted the focus from the traditional certificate functions, we recognize that they still play an important role in verifying control invariant sets. Looking ahead, enhancing the robustness of the RL-DH safety filter against potential distribution shifts will be crucial for the improvement of our RL-based approach. Furthermore, we are keen on extending of our method for uncertain systems, as discussed in \citet{lopez2020robust, brunke2022barrier, cohen2023uncertainty}, which may include extensions to non-affine forms of the control input constraints like second-order cone constraints, as in \citet{castaneda2021pointwise, dhiman2021control, taylor2020towards}, and to multiple constraints, as in \citet{kim2023trust}.

\acks{
This work is supported by the National Science
Foundation Grant CMMI-1944722, the DARPA Assured Autonomy and Assured Neuro Symbolic Learning and Reasoning (ANSR) programs, and the NASA ULI on Safe Aviation Autonomy. The work of Jason J. Choi received the support of a fellowship from Kwanjeong Educational Foundation, Korea. Any opinions, findings, and conclusions or recommendations expressed in this material are those of the authors and do not necessarily reflect
the views of any aforementioned organizations.
}

\section{Appendix}
\label{sec:appendix}

\subsection{Proof of Theorem \ref{thm:main1}}
\label{subsec:proof1}
We prove Theorem \ref{thm:main1} by proving Lemma \ref{cor:theorem1}, since from Definitions \ref{def:cert} and \ref{def:dh}, Lipschitz continuity of $\uref$ implies that $a(x)^\top u \ge b(x)$ is a valid safety constraint, and thus, $\{x | a(x)^\top u = b(x)\}$ being a discriminating hyperplane. Under the conditions in Theorem \ref{thm:main1}, \eqref{eq:dh-qp} can be written as a QP problem which satisfies the conditions in \citet[Thm.1]{morris2013sufficient}, which states that its solution is unique and Lipschitz w.r.t. $x$. \hfill $\square$

\subsection{Proof of Theorem \ref{thm:main2}}
\label{subsec:proof2}

For the proof of the theorem, we consider the distance-like function, $h$, which is continuously differentiable and whose derivative is bounded and Lipschitz continuous, such that $S=\{x\;|\; h(x) \ge 0\}$ and $\dcbf(x)$ is bounded away from 0 for all $x \in \partial S$. As in Lemma \ref{lemma:tangential_ci}, $h$ is solely used as a means to provide proof for the statement. Using $h$, we have $\{u\!\in\!U \;|\;I_S(\traj(\Delta t))\!=\!1\} = \{u\!\in\!U \;|\;h(\traj(\Delta t))\!\ge\!0\}$. From the boundedness of $f, g$ and $U$ from Assumption \ref{assump:dynamics}, we have
\begin{equation*}
|| \traj(\Delta t) - \left(x + \Delta t(f(x) + g(x) u)\right) || \le M_1 \Delta t^2    
\end{equation*}
for all $u \in U$, and for some constant $M_1 > 0$. From the Lipschitz continuity of $\dcbf$, we have
\begin{equation}
\label{eq:lipschitz}
 \dcbf \cdot (y - x) + \frac{L}{2} || y - x ||^2 \ge h(y) - h(x) \ge \dcbf \cdot (y - x) - \frac{L}{2} || y - x ||^2,
\end{equation}
\noindent where $L$ is the Lipschitz continuity of $\dcbf$.
Thus, we have
\begin{equation*}
    M_3 \Delta t^2 + M_4 \Delta t^4 \ge h(\traj(\Delta t)) - h\left(x + \Delta t(f(x) + g(x) u)\right) \ge - M_3 \Delta t^2 - M_4 \Delta t^4
\end{equation*}
\noindent where $M_3 = H M_1$, $M_4 = \frac{1}{2}L M_1^2$, and $H$ is the bound of $||\dcbf||$. Also, with $y=x + \Delta t (f(x) + g(x)u)$ in \eqref{eq:lipschitz}, we have
\vspace{-0.5em}

\small
\begin{equation*}
\dcbf \cdot (f(x) + g(x)u) \Delta t + M_5 \Delta t^2 \ge h\left(x + \Delta t(f(x) + g(x) u)\right) - h(x) \ge \dcbf \cdot (f(x) + g(x)u) \Delta t - M_5 \Delta t^2
\end{equation*}
\vspace{-0.25em}
\normalsize \noindent for some constant $M_5 >0$ for all $u \in U$, due to the boundedness of $f, g$, and $U$. Combining the two above equations, we get for all $u \in U$,
\vspace{-0.75em}
\begin{align}
\label{eq:thm2-2}
\dcbf \cdot (f(x) + g(x)u) & \Delta t \!+\!(M_3 + M_5) \Delta t^2 \!+\!M_4 \Delta t^4 \\
    & \ge
    h(\traj(\Delta t)) - h(x) \ge \dcbf \cdot (f(x) + g(x)u) \Delta t\!-\!(M_3 + M_5) \Delta t^2\!-\!M_4 \Delta t^4. \nonumber
\end{align}
From the right hand side of \eqref{eq:thm2-2}, consider $a(x)^\top\!=\!\dcbf \cdot g(x) \Delta t$, and $b(x)\!=\!-h(x)\!-\!\dcbf \cdot f(x) \Delta t + (M_3 + M_5) \Delta t^2 + M_4 \Delta t^4$, which are both Lipschitz continuous in $x$, due to the Lipschitz continuity of $h$, $\dcbf$, $f$, and $g$. Then $h(\traj(\Delta t)) \ge a(x)^\top u - b(x)$, for all $u\in U$. Thus, if $a(x)^\top u \ge b(x)$, then $h(\traj(\Delta t)) \ge 0$. Therefore, $\Pi(x) = \{u \in U \;|\;a(x)^\top u \geq b(x)\} \subseteq \{u \in U \;|\;I_S(\traj(\Delta t)) = 1\}$. 

Next, from \eqref{eq:thm2-2}, we get 
\vspace{-0.5em} 
\begin{equation*}
    a(x)^\top u - b(x) + 2\left((M_3 + M_5) \Delta t^2 \!+\!M_4 \Delta t^4\right)\ge h(\traj(\Delta t)) \ge a(x)^\top u - b(x).
\end{equation*}
If $\{u \in U \;|\;I_S(\traj(\Delta t)) = 1\}$ has a non-empty interior, $\exists u \in U $ such that $h(\traj(\Delta t)) > 0$, thus, $a(x)^\top u - b(x) + 2\left((M_3 + M_5) \Delta t^2 \!+\!M_4 \Delta t^4\right)> 0$. For small enough $\Delta t$, $a(x)^\top u - b(x) \ge 0$, thus, proving the second statement. \hfill $\square$

\subsection{Implementation Details}
\label{subsec:appendix-implementation}

\textbf{Oracle-DH}: For the Oracle-DH method, we employ Support Vector Machines (SVMs) to determine a discriminating hyperplane at every state along the trajectory. Inputs labeled as `safe' or `unsafe' based on the sample-and-hold approach described in Section \ref{sec:lookahead} are used to train the SVM. The SVM output is then directly utilized as the discriminating hyperplane in the safety filter.
\vspace{1em}

\noindent \textbf{SL-DH}: In the loss function, we typically set $\gamma_{\text{pos}} > \gamma_{\text{neg}}$, since in order for the learned hyperplane to be a valid discriminating hyperplane, it is imperative to not misclassify unsafe inputs. Next, due to prediction error of the neural-network output, the direct use of raw predictions for the safety constraint $a_{\theta}(x)^\top u \geq b_{\theta}(x)$ might not ensure safety. To improve safety, we employ a minor adjustment factor, $\epsilon \geq 0$, to tighten the predicted hyperplane constraint. Specifically, we use the revised constraint, $a_{\theta}(x)^\top u \geq b_{\theta}(x) + \epsilon$. This adjustment can be viewed as a straightforward calibration step, a practice commonly seen in many deep neural network applications \citet{guo2017calibration}. It should be noted, however, that over-utilizing this calibration (i.e., setting an excessively high value for $\epsilon$) can lead to overly conservative behavior by the safety filter. Nevertheless, in our experimental results, we observe no such over-conservatism, indicating minimal, yet effective calibration.
\vspace{1em}

\noindent \textbf{Inverted Pendulum}: The dataset for training SL-DH consists of $10$k states sampled uniformly across $S$, with $300$ inputs sampled uniformly across $U$ per state. The training data is resampled every epoch, where each epoch we take 5 gradient steps. For the labels, we use a lookahead time of $\Delta t=0.05$. We use the following network parameters: $5$ hidden layers, $1000$ hidden layer size, ReLU activations, learning rate of $5\text{e-}4$, $\gamma_{\text{pos}}=10$, $\gamma_{\text{neg}}=1$. We train for $400$ epochs with $5$ gradient steps per epoch. We use $\epsilon = 0.01$ for the calibration.

\vspace{1em}

\noindent \textbf{Kinematic Vehicle:} The system state is defined as $[p_x, p_y, \theta, v]$ which describes the positions, heading with respect to the x-axis, and the velocity of the car. The input is defined as $[\omega, a]$ where $\omega$ is the yaw rate and $a$ is the acceleration of the vehicle. The target constraint set is $X=\{x|\sqrt{p_x^2 + p_y^2} \ge r\}$, where $r=2$. We use input limits $|\omega|\leq 2$, $|a|\leq a_{\text{max}} = 1$. We use the control invariant set $S:=\{x|h(x)\ge0\}$ where $h$ is the CBF designed as:
\begin{equation}
    h(x) = \sqrt{\left(p_x+\frac{v^2}{4a_{\text{max}}}\cos \theta \right)^2 + \left(p_y+\frac{v^2}{4a_{\text{max}}}\sin \theta \right)^2} - \left(r+\frac{v^2}{4a_{\text{max}}}\right)
\end{equation}
The CBF is designed by adding a safety margin to $X^c$, based on the vehicle's current heading angle and the stopping distance, calculated based on the maximum deceleration $a_{\text{max}}$.

The dataset for training SL-DH consists of $8$k states and $500$ inputs per epoch. The system dynamics are discretized under $\Delta t_{\text{sys}}=0.05$ and we compare various lookahead times of the values $\Delta t=0.1, 0.5, 1$. To clearly examine the effect of lookahead time, we keep all training parameters and network parameters unchanged across different lookahead times. We use the following network parameters: $3$ hidden layers, $2000$ hidden layer size, ReLU activations, learning rate of $1$e-$4$, $\gamma_{\text{pos}}=5$, $\gamma_{\text{neg}}=1$. We train for $400$ epochs with $5$ gradient steps per epoch. We use $\epsilon = 0.3$ for the calibration for all lookahead times. Since the heading angle $\theta$ is contained with the values $[0, 2\pi)$, we use $\sin$ and $\cos$ encoding of $\theta$ to preserve this continuity. Specifically, we map the state to the vector $[p_x, p_y, \sin \theta, \cos \theta, v]$ during the forward pass of the hyperplane network.

\vspace{1em}

\begin{figure*}[t!]
    \centering
\adjustbox{width=0.35\columnwidth}{    \includegraphics{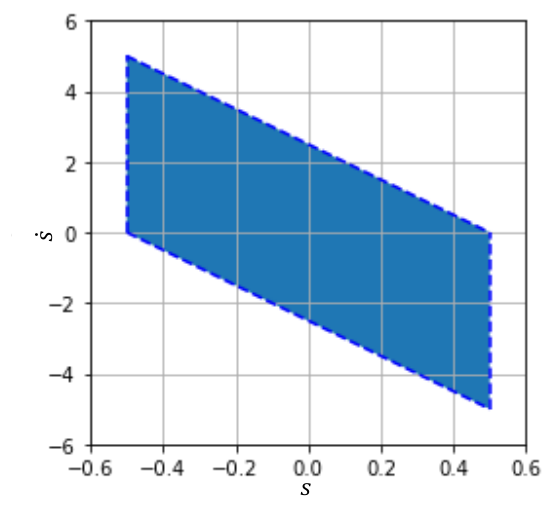}}\vspace{-1em}
    \caption{Hand-designed control invariant set $S$ for the Cart-Pole experiment, used to train SL-DH. The projection of the set to $(s,\dot{s})$ is visualized.} \vspace{-0.5em}
    \label{fig:cart-pole-invariant-set}
\end{figure*}

\noindent \textbf{Cart-Pole:} \textit{(SL-DH training)} For SL-DH, we utilize the geometry of the system to design a control invariant set, visualized in Figure \ref{fig:cart-pole-invariant-set}. We verify this design through a Monte Carlo simulation of trajectories in which we densely sample initial states near the boundary of $S$ and verify based on random trajectories that each state satisfies the control invariance condition. To train SL-DH, we use $10$k steps per epoch with $300$ sampled inputs both uniformly sampled over $S$. We train over $200$ epochs with $5$ gradient steps per epoch. For the labels, we use a lookahead time of $\Delta t_{\text{sys}}$ (1 steps in the system environment). For the network parameters, we use $5$ hidden layers, $1000$ hidden width, ReLU activations, learning rate of $5$e-$4$, $\gamma_{\text{pos}}=5$, $\gamma_{\text{neg}}=1$.

\begin{figure*}[t!]
    \centering
\adjustbox{width=\columnwidth}{    \includegraphics{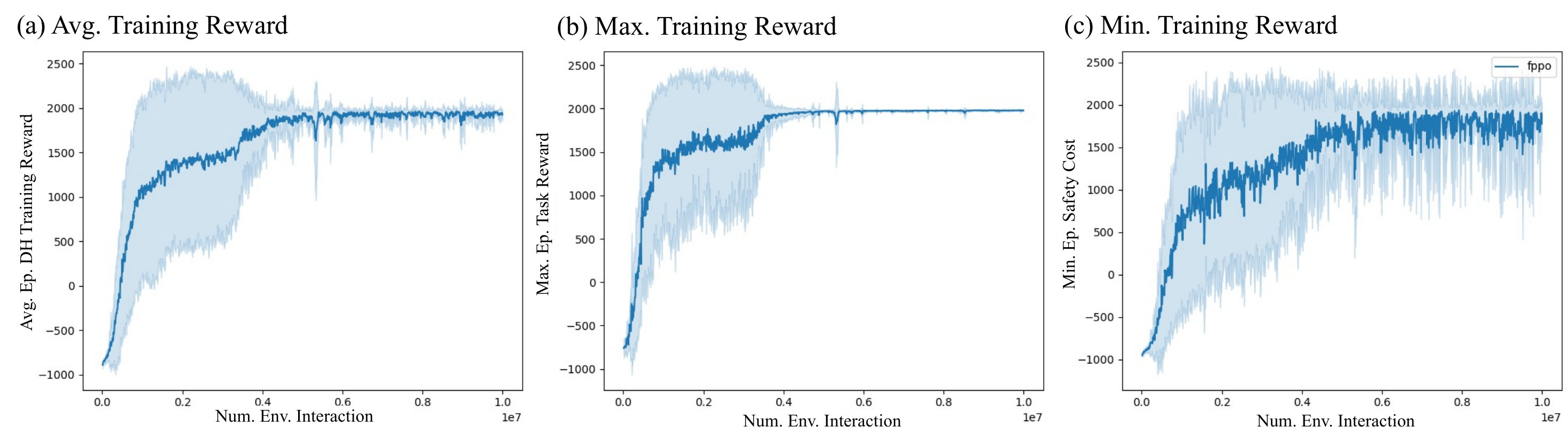}}\vspace{-0.5em}
    \caption{Training curve of RL-DH training for the Cart-Pole Experiment. The reward reaching 2000 indicates that constraint was never violated.}
    \label{fig:cart-pole-rl-dh}
\end{figure*}

\textit{(RL-DH training)} As RL-DH does not rely on a pre-defined safe set, we pretain RL-DH as discussed in Section \ref{sec:rl}, where the environment no longer terminates upon the pole falling. The training curve is reported in Figure \ref{fig:cart-pole-rl-dh}. In detail, our environment uses 1000 steps without termination whereas the classic Cart-Pole environment terminates when the pole falls over, allowing for better exploration in regards to the safety constraints. As detailed in Section \ref{sec:rl}, we use a bonus term $d(u)=1$ when $u$ is inside the action space and $c=-1$ when $x \not \in X$. The training details for RL-DH are as follows: 4000 steps per epoch, 1250 epochs, $\gamma=0.99$, clip ratio of 0.2, actor and critic sizes of 2 hidden layers of 256 width, actor learning rate of $3$e-$4$, critic learning rate of $1$e-$3$, GAE $\lambda$ of 0.97, $80$ gradient steps per actor and critic update steps. 

When utilizing the trained RL-DH for filtering PPO actors for task training, we use a truncated normal distribution whose support region is determined by the discriminating hyperplane to enforce hard safety constraint on the control input. Since $m=1$ for Cart-Pole, we are able to determine the left and right bounds of the truncated normal distribution from the RL-DH hyperplane. Specifically, given the hyperplane parameters $a,b$, the left and right bounds are $[l, r]=[\min(\frac{b}{a}, 1), 1]$ when $a >0$, and $[l, r]=[-1, \max(\frac{b}{a}, -1)]$ when $a < 0$.
We then sample the control input from the truncated normal, $u\sim\mathcal{T}\mathcal{N}(\text{clip}(\mu_{\theta}, l, r), \sigma_{\theta}, l, r)$, where $(\mu_\theta, \sigma_\theta)$ is the mean and variance determined by the actor network, to sample only safe control during rollouts.

\begin{figure*}[t!]
    \centering
\adjustbox{width=\columnwidth}{    \includegraphics{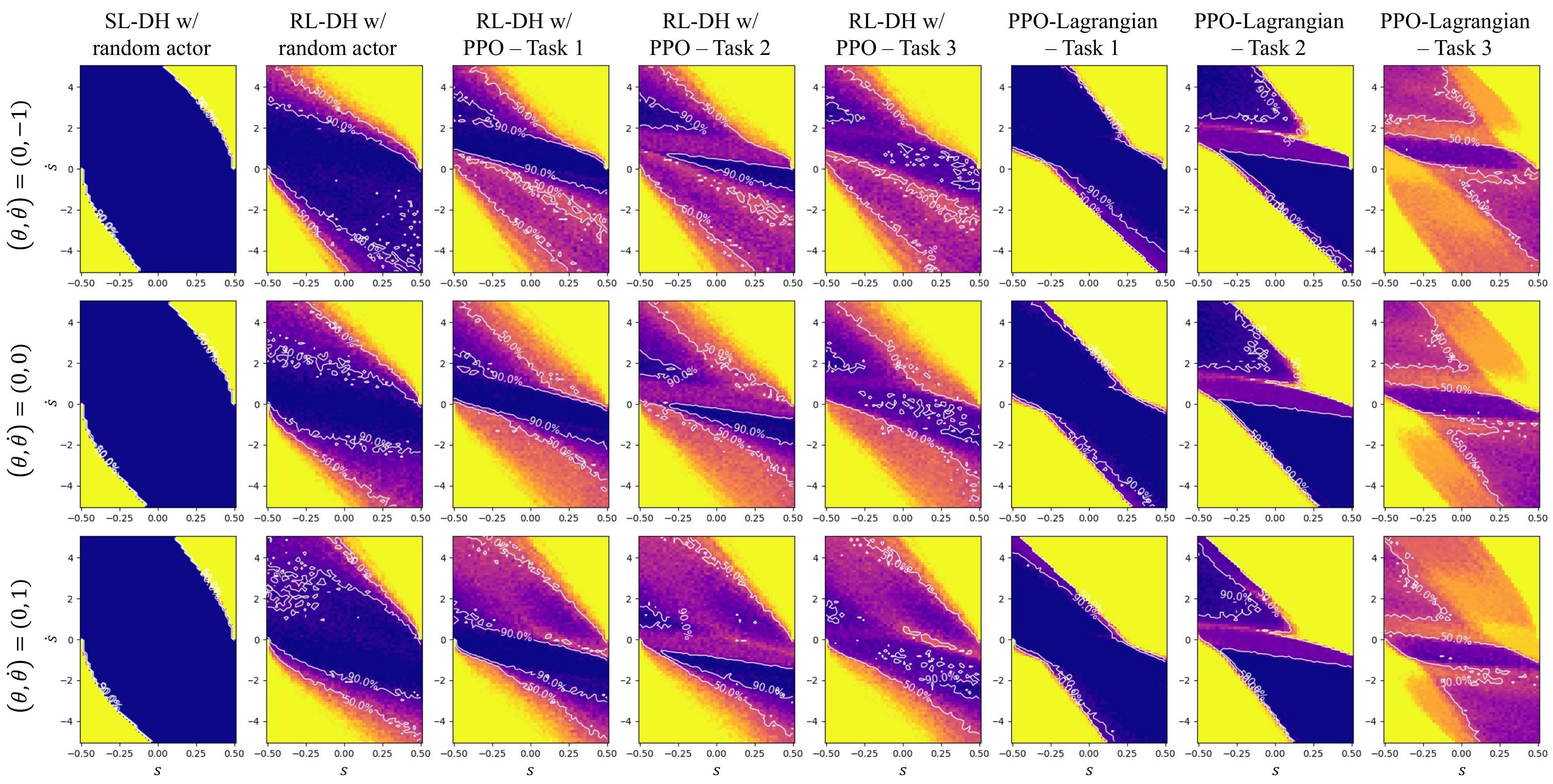}}
    \caption{Constraint satisfaction rates in random rollouts of Cart-Pole trajectories: evaluation at initial states in 2D slices of the state space, at ($\theta, \dot{\theta}$)=(0, -1), (0, 0), (0, 1).}
    \label{fig:cart-pole-safe-sets}
\end{figure*}

\textit{(Safe RL experiments)} The training of PPO, PPO-Lagrangian, PPO with SL-DH, and PPO with RL-DH all use the same parameters as detailed for training the RL-DH, except for the number of epochs. We train each method over 10 different random seeds. 
We train for $62.5$ epochs for PPO, PPO with SL-DH, and PPO with RL-DH, while for PPO-Lagrangian we train for $125$ epochs. We do this to save computation time since non-Lagrangian methods converge to a high-performing policy more promptly than PPO-Lagrangian. For PPO-Lagrangian, we use a learning rate of $5$e-$2$ for the lagrangian multiplier with one gradient step per epoch. The cost function used for PPO-Lagrangian is defined as $c(x)=1$ if $|s|<0.5$, else $c(x)=0$. 

We evaluate all four methods for three different variations of the CartPole task. The first is the classic CartPole objective where the reward is 1 when the pole angle $\theta$ is within $\pm 12$ degrees. The second task consists of a reward of $r(x)=1 + |s-0.4|$. which incentivizes the cart position to be at $s=0.4$. This task is more challenging for the policy to achieve while satisfying the safety constraint since the desired position is close to the safety boundary. The final task uses a reward of $r(s)=1 + |\dot{s}|$. This incentivizes the cart to continue moving as much as possible while keeping the pole upright, making it even harder to satisfy the safety constraint.

Finally, we compare the set of safe initial states across four methods. In addition to the PPO policies filtered by RL-DH safety filters and the PPO-Lagrangian policies for each task, we also include an actor policy initialized as a random policy (with a randomly weighted neural network) that is filtered with SL-DH and RL-DH. This approach effectively isolates the safety-filtered control input from any specific performance objective. For each initial state, selected from grid points within 2D grids in the $(s, \dot{s})$ space, sliced at $(\theta, \dot{\theta})=(0, -1), (0, 0), (0, 1)$, we rollout 50 trajectories (5 trajectories per 10 random seeds) and evaluate whether they exit the set $X$.

We present the success rate of constraint satisfaction for these 50 trajectories in Figure \ref{fig:cart-pole-safe-sets}. Two observations in the figure are particularly noteworthy: First, the SL-DH safety filter provides the largest set of safe initial states among the four methods, with a clear demarcation between safe and unsafe initial states. Second, while the support region of the distribution of safe initial states for PPO-Lagrangian varies with the tasks, the distribution for RL-DH exhibits a similar support region across tasks. This suggests that while PPO-Lagrangian learns task-specific safe policies, the RL-DH-based safety filter is task-agnostic.

\vspace{1em}

\noindent \textbf{HalfCheetah:} The safety constraint we impose is $X=\{x | z \geq -0.3\}$ where z is the height of the robot torso. To train RL-DH, we use a reward of 1 if $x \in X$ and $c=-2$ otherwise. The parameters used to train RL-DH are as follows: 2 hidden layers and 256 hidden width (for both policy and value function networks), discount factor $\gamma=0.99$, $30$k steps per epoch over $167$ epochs, policy learning rate of $3$e-$4$, value learning rate of $1$e-$3$, GAE $\lambda$ of 0.97, and $80$ gradient steps for both the policy and value function.
When using RL-DH as a filter for PPO, we take a different approach than CartPole since it is not possible to come up with a truncated normal distribution form for a higher dimensional control input space. Instead of using a truncated normal distribution, we use a normal distribution but project $\mu_{\theta}$ onto the hyperplane if the constraint $a \mu_{\theta}\geq b$ is not satisfied. Even though this doesn't provide hard safety guarantees for the sampled actions, we found this to provide the best performance while still mitigating safety significantly.

The training for PPO, PPO with RL-DH, and PPO-Lagrangian use the following parameters: 2 hidden layers and 256 hidden width (for both policy and value function networks), discount factor $\gamma=0.99$, $4$k steps per epoch over $250$ epochs, policy learning rate of $3$e-$4$, value learning rate of $1$e-$3$, GAE $\lambda$ of 0.97, and $80$ gradient steps for both the policy and value function. The reported results are from 5 random seeds for each methods.

\bibliography{references}

\end{document}